\documentclass[letterpaper, 10 pt, conference]{ieeeconf}  

\IEEEoverridecommandlockouts                              
\usepackage{graphics} 
\usepackage{epsfig} 
\usepackage{amsmath} 
\usepackage{amssymb}  
\usepackage{mathtools}
\usepackage{algorithmic}
\usepackage{caption}
\usepackage{cite}

\usepackage[ruled,vlined]{algorithm2e}

\DeclareMathOperator*{\argmin}{argmin\,}
\DeclareMathOperator*{\minimize}{minimize\,}

\DeclareMathOperator*{\st}{subject\,to\,}
\DeclareMathOperator*{\sign}{sign}
\DeclareMathOperator\trace{tr}

\newtheorem{theorem}{Theorem}
\newtheorem{corollary}{Corollary}
\newtheorem{lemma}{Lemma}

\title{\LARGE \bf
Robust Online Covariance and Sparse Precision Estimation Under Arbitrary Data Corruption
}

\author{Tong Yao and Shreyas Sundaram
\thanks{Tong Yao and Shreyas Sundaram are with Elmore Family School of Electrical and Computer Engineering at Purdue University. 
Email: \{yao127, sundara2\}@purdue.edu
}  
\thanks{This research was supported by NSF CAREER award 1653648.
}
}

\begin{document}

\maketitle
\thispagestyle{empty}
\pagestyle{empty}

\begin{abstract}
Gaussian graphical models are widely used to represent correlations among entities but remain vulnerable to data corruption. In this work, we introduce a modified trimmed-inner-product algorithm to robustly estimate the covariance in an online scenario even in the presence of arbitrary and adversarial data attacks. At each time step, data points, drawn nominally independently and identically from a multivariate Gaussian distribution, arrive. However, a certain fraction of these points may have been arbitrarily corrupted. We propose an online algorithm to estimate the sparse inverse covariance (i.e., precision) matrix despite this corruption. We provide the error-bound and convergence properties of the estimates to the true precision matrix under our algorithms. 
\end{abstract}

\section{Introduction}
Graph modeling is at the core of modern statistical learning, with applications spanning a wide range of disciplines, including finance and economics \cite{finance}, neuroscience \cite{neuro}, and health and social science \cite{health}.  These problems address the challenges of estimating complex relationships between multiple variables to gain insight into the underlying interactions. 

One way to represent and quantify these relationships is by learning the covariance and inverse covariance matrix through the collected multivariate data, which captures the degree to which the variables change together. Inverse covariance estimation is a crucial task, as the inverse covariance matrix (also known as the precision matrix) can reveal the underlying conditional independence structure between variables. By estimating the sparse inverse covariance matrix, one can identify the most relevant connections between variables, leading to more interpretable and efficient models. 

Assuming that the underlying relationships follow Gaussian distributions, techniques such as graphical lasso \cite{yuan2007model,friedman2008sparse, banerjee2008model} have been developed to tackle the problem by incorporating sparsity-promoting penalties. Different optimization methods have been proposed to solve the graphical lasso problem, including coordinate descent \cite{friedman2008sparse}, proximal methods \cite{rolfs2012iterative,hsieh2013big}, alternating minimization methods \cite{scheinberg2010sparse,dalal2017sparse}, and Newton-conjugate gradient methods \cite{GL-linear}. 

In many real-world scenarios, data is continuously generated and collected, making these traditional batch-processing methods infeasible or computationally expensive. More recently, graphical lasso has been extended to learn the static and dynamic relationships between variables in an online manner \cite{zhou2010time,monti2014estimating,kolar2010estimating,hallac2017network, OGAMA}. Online estimation methods offer several advantages, including scalability, adaptability to changes, 
real-time decision-making, and reduced computational cost.

However, real-world data often contain outliers, corruptions, and even maliciously poisoned data, which severely impact the performance of statistical estimators. Robust estimation aims to develop methods that are less sensitive to such outliers, providing reliable and accurate estimates in the presence of contaminated data. 
Traditional robust mean estimators, such as the median-of-means \cite{ALON1999137} and the trimmed mean \cite{tukey1963trim,ronchetti2009robust} have been explored in the literature. Early works on robust statistics attempt to estimate the mean given an outlier model \cite{tukey1960survey, huber1992robust}, and recent advancements consider stronger contamination models \cite{diakonikolas2020outlier, lugosi2021robust, online_mean}. For a comprehensive overview of robust mean estimation, readers are directed to a survey \cite{lugosi2019survey}. 
Classical robust covariance estimation techniques include Minimum Covariance Determinant \cite{MCD} \cite{fastMCD} to find a given proportion of uncorrupted observations and compute their empirical covariance matrix and the truncated inner product \cite{robust_gaussian} which filters out data with large absolute values to mitigate the impact of arbitrary corruption. These traditional robust estimators were developed for batch datasets and are not suitable for data arriving in a sequence.

Motivated by the streaming nature and the potential corruption of modern datasets, we propose online and robust covariance and sparse inverse covariance estimation algorithms that enable efficient and robust updating of the estimates as new potentially corrupted and noisy data arrive. We show through theoretical performance guarantees and experimental simulations that our algorithms are effective and less sensitive to data corruption. By addressing these problems, we can build more reliable, accurate, and interpretable models, leading to a better understanding of complex systems and more informed decision-making across various domains.

\section{Problem Definition} \label{sec: problem definition}
Consider a set of $p$ random variables $X = \left[\begin{matrix}x_1& x_2 &  \cdots & x_p\end{matrix}\right]^T$, that are jointly Gaussian with zero mean and covariance $S^*$. These variables can be represented by a graph $G = (V, E)$, where $V = \{v_1,\ldots,v_p\}$ is the set of nodes, with each node $v_i$ representing the random variable $x_i$. An edge $(v_i,v_j) \in E$ indicates that variable $x_j$ is conditionally dependent on $x_i$, given all other random variables.  Conversely, if $(v_i,v_j) \notin E$,  $v_j$ is conditionally independent of $v_i$, given all the other variables. This lack of an edge corresponds to a zero-entry in the precision matrix $\theta^* = (S^*)^{-1}$ \cite{scheinberg2010sparse,koller2009probabilistic}.

These relationships (i.e., graph structure) between the variables are unknown {\it a priori}; the goal is to infer the edges of the graph based on samples from random variables.  More specifically, at each time step $t \in \{1,2,\ldots\}$, the network generates a data vector $X_t = [x_{1,t}, x_{2,t}, \ldots, x_{p,t}]^T \in \mathbb{R}^p$. We assume that each $X_t$ is independently and identically sampled from the underlying Gaussian distribution $X_t \sim \mathcal{N}(0, S^*)$.

We consider the existence of an adversary who can inspect all data up to time $t$ and replace them with arbitrary values. 
Let $\mathcal{X} = [X_1, \ldots, X_t] \in \mathbb{R}^{p \times t}$ be the data matrix containing all data points received up to time $t$. 
We say that the dataset $\mathcal{X}$ is $\eta$-corrupted if, for each pair of variables $x_{i}$ and $x_{j}$, where $i,j \in \{1,\ldots,p\}$, the product of their observations $x_{i,1}x_{j,1},\ldots,x_{i,t}x_{j,t}$ contains at most $\eta t$ corrupted entries, where the corruption parameter $\eta \in (0, 1/32)$. We will explain the range of $\eta$ in Section \ref{sec: online}.

\begin{figure}
     \centering
     \includegraphics[width=0.99\linewidth]{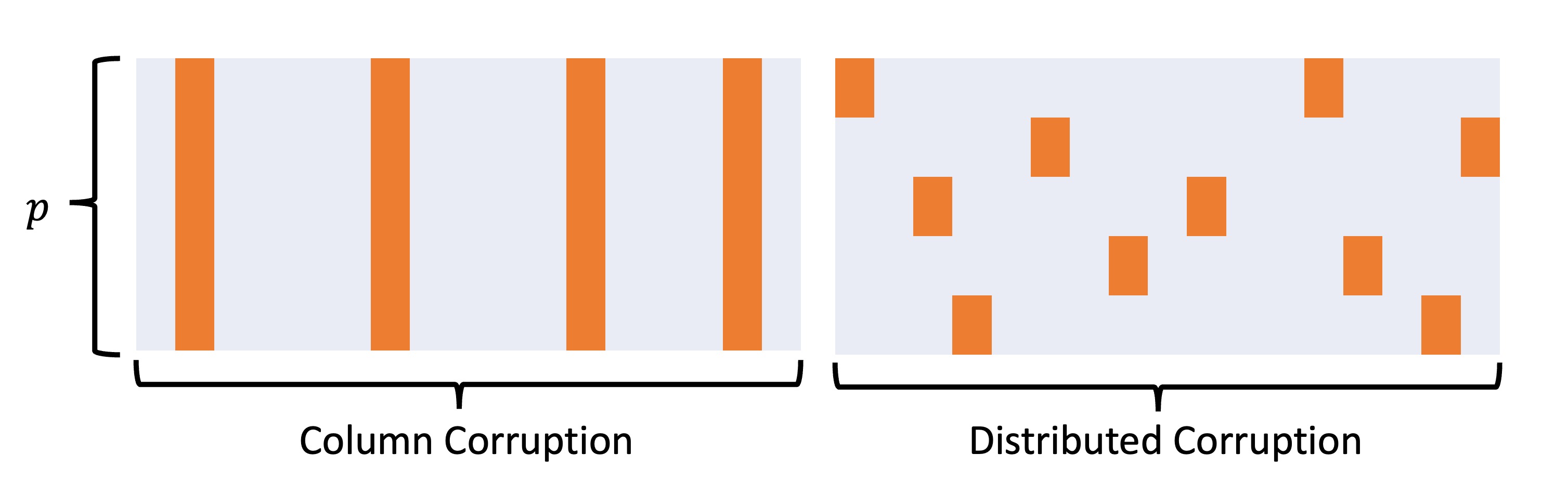}
     \caption{Distribution of corruptions for data matrix, where orange indicates corrupted data points. In these examples, the number of corrupted products of observations is $\eta t = 4.$}
     \label{fig: corruption}
 \end{figure}

In the $\eta$-corruption scenario, the corrupted entries can be located in the same columns (observations) of $\mathcal{X}$ across rows (variables). This \textit{column corruption} allows at most $\eta t$ corrupted data points for each variable.  
The $\eta$-corruption also implies that the corrupted data can be distributed anywhere in $\mathcal{X}$. This is defined as a \textit{distributed corruption} scenario where each variable has at most $\eta/2$ arbitrary corrupted observations.
We show the corruption models in Fig. \ref{fig: corruption}.

We apply the tilde notation to any potentially corrupted data or dataset, e.g., $\tilde{x}_t$ is a potentially corrupted data sample, $\tilde{\mathcal{X}}$ is the $\eta$-corrupted data matrix containing observations up to time $t$, and $\tilde{X}_t = [\tilde{x}_{1,t},\tilde{x}_{2,t},\ldots,\tilde{x}_{p,t}]^T$ is the potentially corrupted observation vector at time $t$.

Given a sequence of potentially corrupted observations $\{\tilde{X}_{1}, \tilde{X}_{2}, \ldots\}$, our objective is to perform real-time (i.e., online) robust estimation of the sample covariance matrix $S^*$ and infer the conditional dependencies between the variables through estimating the sparse precision matrix $\theta^*$.
 
\section{Background} \label{sec: backgroud}
We build our approach on an alternating minimization algorithm proposed in \cite{dalal2017sparse} (batch) and \cite{OGAMA} (online) to solve problem \eqref{eqn: GL}. The online algorithm was shown in \cite{OGAMA} to achieve a similar result with fewer iterations in real-time settings. We provide the background of the algorithm from \cite{OGAMA} and subsequently discuss our modification to account for arbitrary data corruption in Section \ref{sec: online}.

Given a set of uncorrupted data $\{X_1,X_2,\ldots,X_t\}$ up to time $t$, the maximum likelihood estimation problem for recovering $\theta$ is given by
\begin{equation}\label{eqn: GL}
  \minimize_{\theta_t \in \mathcal{S}_{++}^p} -\log\det\theta_t + \trace(S_t\theta_t)+ \lambda|\theta_{t}|_{l1}, 
\end{equation}
where the set of $p \times p$ positive definite matrices is denoted by the set $\mathcal{S}_{++}^p$, and $S_t = \frac{1}{t}\sum_{i=1}^t X_i X_i^T$ is the sample covariance matrix constructed from all the data up to the time step $t$. The terms $-\log\det\theta_t + \trace(S_t\theta_t)$ are derived from the Gaussian log-likelihood function \cite{yuan2007model}, where $\trace$ denotes trace, and the term $|\theta|_{l1} = \sum_{i,j=1}^p |\theta_{ij}|$ is the element-wise $l_1$ norm encouraging sparsity of the solution regulated by the penalty parameter $\lambda \geq 0$.

The approaches in \cite{dalal2017sparse} and \cite{OGAMA} formulate the primal and dual objective functions for problem \eqref{eqn: GL}. The primal of \eqref{eqn: GL} is:
\begin{align} \label{eqn: primal}
\begin{split}
\minimize_{\Omega_t \in \mathcal{S}^p_{++}, \Phi_t \in \mathcal{S}^p_{++}} &-\log\det\Omega_t + \trace(S_t\Phi_t) + \lambda|\Phi_t|_{l1}\\
\st &\Phi_t = \Omega_t.    
\end{split}
\end{align}
The dual of \eqref{eqn: GL} is given by
\begin{align} 
    \minimize_{\Gamma_t\in\mathcal{S}_{++}^p} &-\log\det\Gamma_t - p \label{eqn: dual}\\
    \st &|\Gamma_{ij,t}- S_{ij,t}| \leq \lambda \quad \forall i,j, \nonumber
\end{align}
where the symmetric positive definite matrix $\Gamma_t$ is the dual variable, and $A_{ij,t}$ denotes the $(i,j)$-th element of matrix $A$ at time $t$.

Given the sample covariance matrix $S_t$, the alternating minimization follows the iterative sequence of updates, where each iteration is indexed by the variable $k \in \mathbb{N}$:
\begin{align} 
    \Omega^{k+1}_t & = \argmin_{\Omega \in\mathcal{S}^p_{++}} -\log\det \Omega + \trace(\Gamma_t^k\Omega)\label{eqn: omega update},\\
    \Phi^{k+1}_t &= \argmin_{\Phi\in\mathcal{S}^p_{++}} \trace(S_t\Phi) + \lambda|\Phi|_{l1}-\trace(\Gamma^k_t\Phi)  \nonumber \\
    &\qquad\qquad\qquad\qquad +\frac{\zeta_t^k}{2}\|\Omega_t^{k+1} - \Phi\|_F^2 \label{eqn: phi update},\\
    \Gamma^{k+1}_t &= \Gamma_t^{k}+\zeta_t^k(\Omega_t^{k+1} - \Phi_t^{k+1}) \label{eqn: gamma update}.
\end{align}
In the above equations,  $\zeta_t^k$ is a step size, and $\|A\|_F = \sqrt{\trace(AA^T)}$ denotes the Frobenius norm of a given matrix $A$. The step size $\zeta_t^k$ is chosen to guarantee convergence of the estimates $\Gamma_t^k$ to their desired quantities and can be set as a constant $\zeta = \zeta_t^k < a^2$ for some constant $a$ \cite{OGAMA}. 
The problem of recovering $\theta_t$ reduces to solving \eqref{eqn: omega update}$-$\eqref{eqn: gamma update}. We provide robust estimation details in the next section.

\section{Online Robust Covariance and Precision Estimation} \label{sec: online}
In this section, we will introduce the robust covariance estimator and describe the modification of the sparse precision estimation.

\subsection{Online Trimmed Inner Product}
We describe an online trimmed inner product estimator based on the trimmed mean estimator in \cite{lugosi2021robust} (batch) and \cite{online_mean} (online). We modify the estimator for robust online estimation of covariance. The robust estimator can mitigate the influence of arbitrary corruption and we will show the theoretical performance guarantees in Section \ref{sec: analysis}. 

At the beginning of the algorithm, the system designer will select an initialization time step $t_0 \in \mathbb{N}$, a desired confidence interval $\delta \in (0,1)$, and the corruption level $\eta$. Note that $\eta$ can be an estimate of the upper bound of the corruption rate. 

At each time step, a potentially corrupted data vector $\tilde{X}_t \in \mathbb{R}^p$ arrives. We compute the product $\tilde{s}_t = \tilde{X}_t\tilde{X}_t^T$, where the $(i,j)$-th entry $\tilde{s}_{ij,t} = \tilde{x}_{i,t}\tilde{x}_{j,t}$. 

For $t < t_0$, 
we temporarily store the product matrices in a data vector for each $(i,j)$-th entry $\tilde{Y}_{ij,t} = [\tilde{s}_{ij,1}, \ldots, \tilde{s}_{ij,t}]$.

At $t = t_0$, we start with the set of potentially corrupted products $\tilde{Y}_{ij,t_0} = [\tilde{s}_{ij,1}, \ldots, \tilde{s}_{ij,t_0}]$. 
Given the corruption parameter $\eta$ and the desired confidence level $\delta$, define 
\begin{equation}\label{eqn: epsilon def}
\epsilon = 8\eta + 12\frac{\log(4/\delta)}{t_0}.
\end{equation}
Note that for sufficiently large $t$, $\epsilon < 0.25$ since $\eta < 1/32$. 
For each $\tilde{Y}_{ij,t_0}$, let $\tilde{s}_{ij,1}^* \leq \tilde{s}_{ij,2}^* \leq \ldots \leq \tilde{s}_{ij,t_0}^*$ be a non-decreasing rearrangement of $\tilde{s}_{ij,1}, \ldots, \tilde{s}_{ij,t_0}$. 
For each $i,j \in \{1,\ldots, p\}$, define $\alpha_{ij} = \tilde{s}_{ij, \epsilon t}^*$ and $\beta_t = \tilde{s}_{ij,(1-\epsilon)t}^*$ to be {\it trimming values}. 
For each $\alpha_{ij} \leq \beta_{ij}$, and $s \in \mathbb{R}$, define the \textit{trim estimator} 
\begin{equation}
\phi_{\alpha_{ij},\beta_{ij}}(s) = 
\begin{cases}
\beta_{ij} &\text{if } s > \beta_{ij},\\
s &\text{if } s \in [\alpha_{ij},\beta_{ij}],\\
\alpha_{ij} &\text{if } s < \alpha_{ij}.
\end{cases}    
\end{equation}
We then initialize the $(i,j)$-th entry of the robust estimation of sample covariance 
\begin{equation}\label{eqn: robust cov init}
    \hat{S}_{ij,t_0} = \frac{\sum_{k=1}^{t_0} \phi_{\alpha_{ij},\beta_{ij}}(\tilde{s}_{ij,k})}{t_0}.
\end{equation}

For all $t > t_0$, the process recursively updates the estimate using the previous estimate $\hat{S}_{ij,t-1}$ and the new product $\tilde{s}_{ij,t}$ 
\begin{equation}\label{eqn: robust cov update}
    \hat{S}_{ij,t} = \frac{(t-1)\hat{S}_{ij,t-1} + \phi_{\alpha_{ij},\beta_{ij}}(\tilde{s}_{ij,t})}{t}.
\end{equation}

We include the pseudo-code implementation for the online robust covariance estimation in Algorithm \ref{algo: trim}. 

\begin{algorithm}
\SetKwInOut{Parameter}{Parameter}
\Parameter{$t_0$, $\delta$, $\eta$}
\KwIn{$\tilde{X}_t = [\tilde{x}_{1,t}, \ldots, \tilde{x}_{p,t}]$}
\For{$i, j \in \{1, 2, \ldots, p\}$}{
Compute $\tilde{s}_{ij,t} = \tilde{x}_{i,t} \tilde{x}_{j,t}$\\
\If{$t<t_0$}{
Store data in $\tilde{Y}_{ij,t_0}$
}
\If{$t = t_0$}{
Compute $\epsilon = 8\eta + 12\frac{\log(4/\delta)}{t_0}$\\
Sort data in $\tilde{Y}_{ij,t_0}$ and determine trimming thresholds $\alpha_{ij}, \beta_{ij}$\\
$\hat{S}_{ij,t_0} = \frac{\sum_{k=1}^{t_0} \phi_{\alpha_{ij},\beta_{ij}}(\tilde{s}_{ij,k})}{t_0}$
}
\If{$t>t_0$}{
update trimmed inner product $\hat{S}_{ij,t} = \frac{(t-1)\hat{S}_{ij,t-1} + \phi_{\alpha_{ij},\beta_{ij}}(\tilde{s}_{ij,t})}{t}$
}}
\Return $\hat{S}_t$
\caption{Online Trimmed Inner Product}
\label{algo: trim}
\end{algorithm}

\begin{algorithm}
\SetKwInOut{Parameter}{Parameter}
\Parameter{$\lambda$, $t_0$}
\KwIn{Stream of potentially corrupted multivariate data $\tilde{X}_1, \tilde{X}_2, \ldots \in \mathbb{R}^{p}$}

\For{$t\in \{1, 2, \ldots\}$}{
$\hat{S}_t = $ online-trimmed-inner-product($\tilde{X}_t$)\\
\If{$t=t_0$}{
$\Gamma_{t_0} = \hat{S}_{t_0}+\lambda I_p$\\
Choose $\zeta_t \in (0, (\lambda_{\min}(\Gamma_{t_0}))^2)$\\
}
\ElseIf{$t>t_0$}{
$\Gamma_{t}= \mathcal{C}_\lambda(\Gamma_t-\hat{S}_t+\zeta_{t-1}(\Gamma_{t-1})^{-1})+\hat{S}_t$\\
Choose $\zeta_{t} \in (0, (\lambda_{\min}(\Gamma_t))^2)$

Update $\Omega_t$ and $\Phi_t$ as per \eqref{eqn: omega} and \eqref{eqn: phi}
}
}
\Return Sequence of sparse precision estimates $\Phi_{t_0+1}, \Phi_{t_0+2},\ldots$
\caption{Online Graphical Alternating Minimization Algorithm (O-GAMA)}
\label{algo: OGAMA}
\end{algorithm}

\subsection{Online Alternating Minimization Algorithm}\label{sec: OGAMA}

At the initialization time step $t_0$, we obtain $\hat{S}_{t_0}$ from \eqref{eqn: robust cov init} and initialize $\Gamma_{t_0} = \hat{S}_{t_0} + \lambda I_p$, where $I_p \in \mathbb{R}^{p \times p}$ is an identity matrix, with data up to a user-defined $t_0 \in \{1,2,\ldots\}$. The sparsity penalty parameter $\lambda$ ensures that the initial dual variable is positive definite, i.e., $\lambda > \max\{0, -\lambda_{\min}(\hat{S}_{t_0})\}$.

We apply the robust covariance estimator and simplify the alternating minimization algorithm by allowing one optimization iteration per time step. Taking the derivatives of the expressions for $\Omega_t$ and $\Phi_t$ and equating them to 0,  we obtain the closed-form updates of \eqref{eqn: omega update} and \eqref{eqn: phi update}:
\begin{align} 
    \Omega_t &= (\Gamma_{t})^{-1},\label{eqn: omega}\\
    \Phi_t &= \frac{1}{\zeta_{t-1}}\mathcal{S}_\lambda(\zeta_{t-1}\Omega_t - \hat{S}_t + \Gamma_{t}) \label{eqn: phi}.
\end{align}
Here, $\mathcal{S}_\lambda(x) = \sign(x)(\max(|x|-\lambda,0))$ is the soft-thresholding operator (applied element-wise to a matrix argument). Following these update rules, $\Omega_t$ is interpreted as an approximately sparse precision matrix and $\Phi_t$ is interpreted as the estimate of the sparse precision matrix. Substituting \eqref{eqn: omega} and \eqref{eqn: phi} for the variables in \eqref{eqn: gamma update}, and using the clip function $\mathcal{C}_\lambda(x) = \min(\max(x,-\lambda),\lambda)$ with property $x = \mathcal{S}_\lambda(x) + \mathcal{C}_\lambda(x)$, the dual update \eqref{eqn: gamma update} can be written as:
\begin{align} \label{eqn: dual update}
    \Gamma_t = \mathcal{C}_\lambda(\Gamma_{t-1} - \hat{S}_t + \zeta_{t-1}(\Gamma_{t-1})^{-1})+\hat{S}_t.
\end{align}
The algorithm first iterates through \eqref{eqn: dual update} to obtain the new dual variable. The updated dual variable is then used to update \eqref{eqn: omega} and \eqref{eqn: phi}.

We provide the pseudo-code implementation of the robust sparse precision estimation algorithm in Algorithm \ref{algo: OGAMA}.

\section{Theoretical Analysis}\label{sec: analysis}
In this section, we provide theoretical guarantees of the quality of the robust sample covariance estimator and then analyze the quality of the sparse precision estimation.
\subsection{Robust Sample Covariance}
Let $tS = \sum_{k = 1}^t X_kX_k^T$ be a $p \times p$ random matrix, where each $X_t$ is i.i.d. sampled from $\mathcal{N}(0,S^*)$. The matrix $tS$ follows the Wishart distribution $tS \sim \mathcal{W}(S^*, t)$ which arises as the distribution of the sample covariance matrix for a sample of a multivariate normal distribution \cite{wishart}. 
Using results from the Wishart distribution, each $(i,j)$-th entry of $S$, denoted $S_{ij}$, is a real-valued random variable with mean $S^*_{ij}$ and finite variance $\sigma^2_{S_{ij}} ={S_{ij}^*}^2 + {S}^*_{ii}{S}^*_{jj}$.
  
We set up the analysis for each $(i,j)$-th entry of $S$. For simplicity, we omit the subscript $ij$ of $S$ for the following analysis. Define $\overline{S} = S-S^*$.
For $0<q<1$, define the quantile 
\begin{equation}\label{eqn: def Q}
Q_q(\overline{S}) = \sup\{M \in \mathbb{R}: \mathbb{P}(\overline{S} \geq M) \geq 1-q\}.    
\end{equation} We have $\mathbb{P}(\overline{S} \geq Q_q(\overline{S})) = 1-q$ and from Chebyshev’s inequality, 
\begin{equation*}
1-q = \mathbb{P}(\overline{S} \geq Q_q(\overline{S})) \leq \mathbb{P}(|\overline{S}| \geq Q_q(\overline{S}))\leq \frac{\sigma_S^2}{ Q_q^2(\overline{S})}.   	
\end{equation*}
As a result,
\begin{equation}\label{eqn: upper bound on Q}
|Q_{q}(\overline{S})| \leq \frac{\sigma_S}{\sqrt{1-q}}.
\end{equation}

The following result upper bounds the trimming values using quantiles. 

\begin{lemma}[\cite{lugosi2021robust}]\label{lemma: Q bounds}
Consider the corruption-free samples $y_1,\ldots,y_t$. With probability at least $1-4e^{-\epsilon t/12}$, the inequalities
\begin{align*}
    |\{i: y_i \geq S^* + Q_{1-2\epsilon}(\overline{S}\}| &\geq 3/2\epsilon t\\
    |\{i: y_i \leq S^* + Q_{1-\epsilon/2}(\overline{S}\}| &\geq (1-(3/4)\epsilon)t\\
    |\{i: y_i \leq S^* + Q_{2\epsilon}(\overline{S}\}| &\geq 3/2\epsilon t\\
    |\{i: y_i \geq S^* + Q_{\epsilon/2}(\overline{S}\}| &\geq (1-(3/4)\epsilon)t
\end{align*} hold simultaneously. We denote the event when the above four inequalities hold as event $A$. On event $A$, since corruption $\eta \leq \epsilon/8$, the following inequalities also hold 
\begin{equation}\label{eqn: beta}
    Q_{1-2\epsilon}(\overline{S}) \leq \beta - S^* \leq Q_{1-\epsilon/2}(\overline{S}),
\end{equation}
\begin{equation} \label{eqn: alpha}
    Q_{\epsilon/2}(\overline{S}) \leq \alpha - S^*\leq Q_{2\epsilon}(\overline{S}).
\end{equation} 
\end{lemma}

With the background mentioned above, we provide the following result on the error of the robust estimation of covariance. Note that the proofs of all results are included in the Appendix.

\begin{theorem}\label{thm: online 1}
Let $t_0 > \max(\frac{3\log(8/\delta)}{2\eta}, \frac{12\log(4/\delta)}{0.25 - 8\eta})$ and fix $\delta \in [4e^{-t_0},1)$.
Following the procedures of Algorithm \ref{algo: trim}, with probability at least $1-\delta$, the estimates satisfy
\begin{multline}\label{eqn: mu bound algo 1}
|\hat{S}_{ij,t}  - S^*_{ij}| \leq \left(\sqrt{2} + \frac{\sqrt{6}}{9}\right) \sigma_{S_{ij}} \sqrt{\frac{\log(4/\delta)}{t}} \\
+ \frac{43\sqrt{2}}{12} \sigma_{S_{ij}}\sqrt{\epsilon}, \forall t \geq t_0,
\end{multline} where $\sigma_{S_{ij}} = \sqrt{{S_{ij}^*}^2 + {S}^*_{ii}{S}^*_{jj}}$.
\end{theorem}

The theorem illustrates that for each $(i,j)$-th entry of the sample covariance, the estimation error consists of an initialization error derived from setting the trimming thresholds and a convergence error influenced by the variance of $x_i, x_j$, as well as the covariance between these variables. Intuitively, robust covariance estimation becomes more challenging in the presence of corruption when $x_i$ and $x_j$ exhibit large variances, or when their covariance is large, implying considerable joint variation.

We provide the following result on the convergence of the robustly estimated sample covariance matrix. 
\begin{corollary}\label{cor: sample cov asymp bound}
Let $t_0 > \max(\frac{3\log(8/\delta)}{2\eta}, \frac{12\log(4/\delta)}{0.25 - 8\eta})$ and fix $\delta \in [4e^{-t_0},1)$.
Following the procedures of Algorithm \ref{algo: trim}, there exists a set of sample paths of measure 1, such that for each sample path in that set, there exists a finite time $\bar{t}$, such that for all $t\geq \bar{t}$, the robust sample covariance satisfies 
\begin{multline}
\|\hat{S}_{t}  - S^*\|_F \leq \left(\sqrt{2} + \frac{\sqrt{6}}{9}\right)p \sigma_{S} \sqrt{\frac{\log(4/\delta)}{t}} \\
+ \frac{43\sqrt{2}}{12} p \sigma_{S}\sqrt{\epsilon}, \forall t \geq t_0.
\end{multline} 
where $\sigma_{S} = \max_{ij}\sqrt{{S_{ij}^*}^2 + {S}^*_{ii}{S}^*_{jj}}$.
\end{corollary}

Using the definition of $\epsilon$ in \eqref{eqn: epsilon def} with Corollary \ref{cor: sample cov asymp bound}, we obtain the result below on the convergence of estimates.
\begin{corollary}\label{coro: asymp cov}
Let $t_0 > \max(\frac{3\log(8/\delta)}{2\eta}, \frac{12\log(4/\delta)}{0.25 - 8\eta})$ and fix $\delta \in [4e^{-t_0},1)$. Following the procedures of Algorithm \ref{algo: trim}, the robust estimate of sample covariance satisfies the following inequality almost surely, 
\begin{equation}
\limsup_{t\to\infty}\|\hat{S}_{t}  - S^*\|_F \leq \frac{43}{6}p\sigma_{S}\sqrt{4\eta + 6\frac{\log(4/\delta)}{t_0}},
\end{equation} 
where $\sigma_{S} = \max_{ij}\sqrt{{S_{ij}^*}^2 + {S}^*_{ii}{S}^*_{jj}}$.
\end{corollary}

From the above results, we observe that the estimation error converges to the error introduced by initialization. Thus, it requires a relatively large $t_0$ to give reasonable estimates.

\subsection{Robust Sparse Precision Matrix}
In this subsection, we provide error bounds for the estimates of sparse precision estimates. It was shown in \cite{dalal2017sparse}, that $\Omega^*=\theta^*$ is the optimal solution of \eqref{eqn: GL} given $S^*$, if and only if $\Gamma^* = (\Omega^*)^{-1}$ is a fixed point of \eqref{eqn: dual update} given $S^*$ (i.e., with access to the true covariance matrix). We will analyze the dual variable $\Gamma_t$ given $t$ data points and as the number of data points increases.

We provide the following theorem to show the lower and upper bound of eigenvalues of the dual variables $\Gamma_t$.
The robustly estimated sample covariance matrix is real and symmetric. However, $S_t$ may no longer be positive semidefinite. For the inverse covariance estimation, we will use the penalty parameter $\lambda$ to enforce positive definiteness in the estimates. 

\begin{theorem}\label{theorem: eigen}
Let $t_0 > \max(\frac{3\log(8/\delta)}{2\eta}, \frac{12\log(4/\delta)}{0.25 - 8\eta})$ and fix $\delta \in [4e^{-t_0},1)$. Define $b = \max_{t \geq t_0}\lambda_{\max}(\hat{S}_t)+p\lambda$ and $a =\min_{t \geq t_0} e^{g(t)}b^{1-p}$, where $g(t) = \log \det \Gamma_{t_0}  - \sum_{k =t_0}^{t-1} \Delta_k$, and $\Delta_t = \trace\left(\left(\Gamma_{t+1}-\Gamma_t\right)(-\Omega_{t})\right)+\frac{1}{2\zeta_t}\|\Gamma_{t+1}-\Gamma_t\|_F^2.$ For any finite $t \ge t_0$, there exists a sufficiently large $\lambda$ such that iterates of $\Gamma_t$ of Algorithm \ref{algo: OGAMA} satisfy $0 \prec aI_p \preceq \Gamma_t \preceq bI_p$.
\end{theorem}

Note that here we provide the result of boundedness for finite time steps, and we can always select a sufficiently large penalty parameter $\lambda$ such that estimates are bounded for finite time steps. We note that $\hat{S}_t$ converges to a range of $S^*$ almost surely and $\Gamma_{t}$ converges to a range of $\Gamma^*$ almost surely. The upper bound $b$ is finite almost surely.  For the lower bound, we observe from experiments that $\sum_{k = t_0}^t {\Delta_k}$ is bounded for all $t \geq t_0$ (see Fig. \ref{fig: error sum} in the Appendix). As a result, the appropriate value for the penalty parameter is also bounded, i.e., $\lambda \in (0,\infty)$. 

To analyze the finite-time performance of robust sparse precision estimation, we introduce the following theorem. The proof is similar to the online graphical alternating minimization algorithm \cite{OGAMA}, and we incorporate the robust sample covariance estimation error.
\begin{theorem}\label{theorem1}
Let $t_0 > \max(\frac{3\log(8/\delta)}{2\eta}, \frac{12\log(4/\delta)}{0.25 - 8\eta})$ and fix $\delta \in [4e^{-t_0},1)$. Assume $\forall t \geq t_0$, the quantities $\Gamma_t$ satisfy $aI\preceq \Gamma_t \preceq bI$ and $\zeta_t = \zeta$, and assume $\Gamma^*$ satisfies $a I_p\preceq \Gamma^* \preceq b I_p$. Then at time-step $t+1$, the dual variable $\Gamma_{t+1}$ satisfies the following condition: 
    \begin{multline}
    \|\Gamma_{t+1}-\Gamma^*\|_F 
    \leq r^{t+1-t_0}\|\Gamma_{t_0}-\Gamma^*\|_F \\
  +2\sum_{k=t_0}^{t} r^{t-k}\|\hat{S}_{k+1}-S^*\|_F,
    \label{eqn:bound_gamma_t}    
    \end{multline}
where $r = \max\left\{|1-\frac{\zeta}{a^2}|,|1-\frac{\zeta}{b^2}|\right\}.$
\end{theorem}

We defer discussing the implications of the result until the asymptotic analysis is presented later in this section.

For analysis, we require the following lemma.
\begin{lemma}\label{lemma:nedic}
Let $0<r<1$ and let $\{\rho_t\}$ be a positive scalar sequence. Assume that $\lim_{t\to\infty}\rho_t \leq \bar{\rho}$. Then
\begin{equation}
  \limsup_{t\to \infty}\sum_{l=0}^{t} r^{t-l}\rho_l \leq \frac{\bar{\rho}}{1-r}. 
\end{equation}
\end{lemma}

\begin{corollary}\label{cor: convergence}
Let $t_0 > \max(\frac{3\log(8/\delta)}{2\eta}, \frac{12\log(4/\delta)}{0.25 - 8\eta})$ and fix $\delta \in [4e^{-t_0},1)$. Assume that there exist constants $0 < a < b$ such that, for all $t \ge t_0$, the quantities $\Gamma_t$ computed in Algorithm \ref{algo: OGAMA} satisfy $aI\preceq \Gamma_t \preceq bI$ and $\zeta_t = \zeta < a^2$. 
Then, as the number of data points $t \to \infty$, the result $\Gamma_{t}$ converges to a range of the optimal solution $\Gamma^*$ almost surely:
\begin{equation}
    \limsup_{t\to \infty}\|\Gamma_{t+1} - \Gamma^*\|_F \leq \frac{43}{6(1-r)}\sigma_S\sqrt{4\eta + 6\frac{\log(4/\delta)}{t_0}},
\end{equation} 
where $r = \max\left\{|1-\frac{\zeta}{a^2}|,|1-\frac{\zeta}{b^2}|\right\}$ and $\sigma_{S} = \max_{ij}\sqrt{{S_{ij}^*}^2 + {S}^*_{ii}{S}^*_{jj}}$.
\end{corollary}

From the above results, the estimation error is introduced by the initialization of the dual variable and the initialization of the trimming threshold. Since the error introduced by dual initialization converges exponentially to 0 (for $0<r<1$), the convergence of the dual variable is dominated by the convergence of the sample covariance matrices. Asymptotically, the estimation error is introduced only by the trimming threshold determined during the initialization process. Notably, the error correlates with the corruption rate $\eta$ but not to any severity of the corruption.

\section{Experiments}
In this section, we demonstrate the effectiveness of robust online estimation algorithms through experimental simulations.
\subsection*{Experimental Setup}
We first generate a sparse Erdos-Renyi network with $p$ nodes following the steps in \cite{gengraph}. We generate a $p \times p$ symmetric matrix $A$ by setting the probability of two nodes having no edge as $0.95$, and the probability of two nodes having an edge as $0.05$; in the latter case, the value of the corresponding entry is chosen to be uniformly distributed within certain intervals: 
\begin{equation*}
    A_{ij}=\begin{cases} 0 & \text{Pr} = 0.95 \\
\text{Unif}([-0.6,-0.3] \cup [0.3,0.6]) &\text{otherwise}
\end{cases}.
\end{equation*}
To ensure that the covariance matrix is positive definite, we set $\theta^* = {(S^*)}^{-1} = A + (\xi + |\lambda_{\min}(A)|)I$, and adjust $\xi$ to make $\lambda_{\min}(\theta^*) = 1$. 

We then generate a clean data matrix $X\in\mathbb{R}^{p\times t}$ from a Gaussian distribution $\mathcal{N}(0, \theta^*)$. For each row in the data matrix, we randomly select $\eta t$ of them to be corrupted, where $\eta = 0.03$. The corruption data are generated by two normal distributions $\mathcal{N}(\mu,\sigma)$, representing small and large corruptions, respectively, as follows:

$$(1)\, \mu = 1, \sigma = 2; \qquad (2)\, \mu = 1, \sigma = 5.$$

\subsection*{Performance of Covariance and Sparse Precision Estimates}
\begin{figure}
    \centering
    \includegraphics[width = 0.9\linewidth]{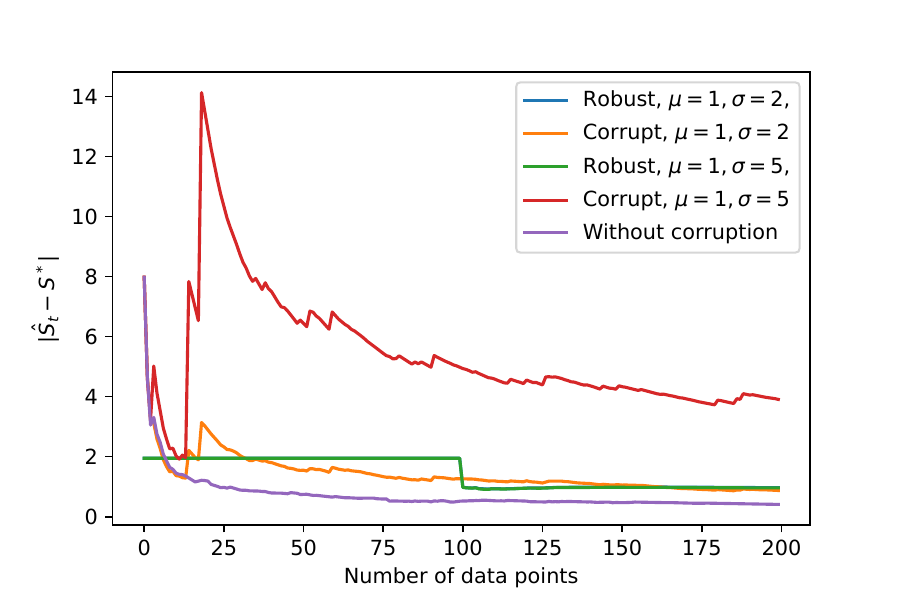}
    \caption{Estimate of sample covariances. }
    \label{fig: cov}
\end{figure}

\begin{figure}
    \centering
    \includegraphics[width = 0.9\linewidth]{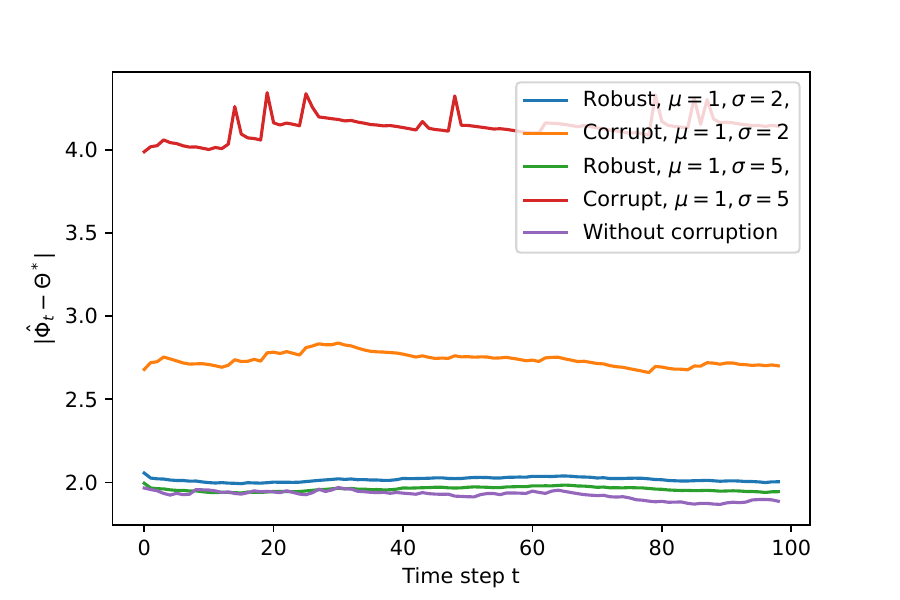}
    \caption{Estimate of sparse inverse covariances.}
    \label{fig: pre}
\end{figure}

For this set of experiments, we let $p = 10, t_0 = 100, \delta = 0.9$, and $\lambda = 0.15$.
We include the simulation results for robust covariance estimation in Fig. \ref{fig: cov} and robust sparse precision estimation in Fig. \ref{fig: pre}.

In Fig. \ref{fig: cov}, we let $\hat{S}_t$ be zero matrices for all $t \leq t_0$. The performance of the two robust estimators is similar and the deviation curves overlap in the plot. We observe that the robust covariance estimator is effective against large corruption and shows improvement over small corruption. Asymptotically, the performance of the robust estimator is influenced by the initialization error and stays within a bounded range of the true covariance.

In Fig. \ref{fig: pre}, we observe that both small and large corruptions have a significant effect on the estimates of the sparse precision matrix. However, with the proposed robust estimator, the estimates are significantly improved and are bounded within a close range of the estimates without any corruption. 

\section{Conclusion and Future Work}
In this work, we proposed online robust covariance and sparse precision estimators. In \cite{online_mean}, it was shown that by trading off computation and memory complexity, the initial estimation error can be eliminated. We will include such modifications to reduce initialization errors in future work.

\bibliographystyle{unsrt}
\bibliography{sample}

\newpage
\appendix
\subsection{Proof of Theorem \ref{thm: online 1}}
First, we state the following lemma to complete the proof.
\begin{lemma}\label{lemma: bern}(Bernstein's inequality)
Let $S_1, \ldots, S_t$ be independent random variables, such that $|S_i- \mathbb{E}[S_i]| \leq a$ for all $i$. Then for any $\delta \in (0,1)$ and $t \in \mathbb{N}$, we have
\begin{multline}\label{eqn: bernstein}
\mathbb{P}\Big(\left|\frac{1}{t}\sum_{k=1}^t (S_k - \mathbb{E}[S_k])\right| \leq \frac{2a}{3t}\log(2/\delta)+\sqrt{\frac{2\sigma^2\log(2/\delta)}{t}} \Big) \\ \geq 1 - \delta, 
\end{multline} where $\sigma^2 = \frac{1}{t}\sum_{k=1}^t \sigma_{S_k}^2$.
\end{lemma}

To simplify notation, we omit the subscript $ij$ that denotes the $(i,j)$-th entry in the proof and let $\phi_{t_0} = \phi_{\alpha_{t_0},\beta_{t_0}}$. From the triangle inequality, we can separate 
\begin{multline}\label{eqn: estimate and true 1}
    \left|\frac{1}{t}\sum_{k = 1}^t \phi_{t_0}(\tilde{s}_k)- S^* \right| 
    \leq \left|\mathbb{E}[\phi_{t_0}(S)] - S^* \right| \\
    + \left|\frac{1}{t}\sum_{k = 1}^t \phi_{t_0}({s}_k)-\mathbb{E}[\phi_{t_0}(S)]\right| \\
    + \Big|\frac{1}{t}\sum_{k = 1}^t \phi_{t_0}(\tilde{s}_k)-\frac{1}{t}\sum_{k = 1}^t \phi_{t_0}(s_k)\Big| .
\end{multline}
The error between the expected values of the trimming operator and the true mean (i.e., $S^* = \mathbb{E}[S]$) is introduced by data beyond the trimming thresholds
\begin{multline} \label{eqn: thm1 bias 1}
    \left|\mathbb{E}[\phi_{t_0}(S)] - S^* \right| \leq |\mathbb{E}[(\alpha_{t_0} - S) \mathbf{1}_{S \leq \alpha_{t_0}}] 
    \\ + \mathbb{E}[(S - S) \mathbf{1}_{{\alpha}_{t_0} < S < {\beta}_{t_0}}]
     + \mathbb{E}[(\beta_{t_0}-S) \mathbf{1}_{S \geq \beta_{t_0}}] |.
\end{multline}
Notice that on the right-hand side of \eqref{eqn: thm1 bias 1}, the first term is positive, the second term is zero, and the third term is negative. 

For $t_0 \geq \frac{3\log(8/\delta)}{2\eta}$, the probability of event $A$ in Lemma \ref{lemma: Q bounds} satisfies $4e^{-\epsilon t_0/12} \leq 4e^{-\frac{2}{3}\eta t_0} \leq {\delta}/{2}$. With probability at least $1-\delta/2$, on event $A$, the trimming thresholds satisfy \eqref{eqn: beta} and \eqref{eqn: alpha}, and recall $\overline{S} = S - S^*$,
\begin{multline} \label{eqn: bias 1}
    \left|\mathbb{E}[\phi_{t_0}(S)] - S^* \right| \\
    \leq \max \Big\{\left|\mathbb{E}[(Q_{2\epsilon}(\overline{S}) - \overline{S}) \mathbf{1}_{\overline{S} \leq Q_{2\epsilon}(\overline{S})}]\right|, \\
    \left|\mathbb{E}[(\overline{S} - Q_{1-2\epsilon}(\overline{S})) \mathbf{1}_{\overline{S} \geq Q_{1-2\epsilon}(\overline{S})}]\right|\Big\}.
\end{multline}

The first term on the right-hand side of \eqref{eqn: bias 1} can be upper bounded as
\begin{align*}
&|\mathbb{E}[(Q_{2\epsilon}(\overline{S}) - \overline{S}) \mathbf{1}_{\overline{S} \leq Q_{2\epsilon}(\overline{S})}]| \\
&\leq |\mathbb{E}[Q_{2\epsilon}(\overline{S})\mathbf{1}_{\overline{S} \leq Q_{2\epsilon}(\overline{S})}]| + |\mathbb{E}[\overline{S}\mathbf{1}_{\overline{S} \leq Q_{2\epsilon}(\overline{S})}]| \\
&\overset{(a)}\leq |Q_{2\epsilon}(\overline{S})|\mathbb{P}(\overline{S} \leq Q_{2\epsilon}(\overline{S})) \\
&\qquad + \sqrt{\mathbb{E}[\overline{S}^2]\mathbb{E}[(\mathbf{1}_{\overline{S} \leq Q_{2\epsilon}(\overline{S})})^2]}\\ 
&\overset{(b)}\leq \frac{\sigma_S}{\sqrt{1- 2\epsilon}}2\epsilon + \sigma_S\sqrt{\mathbb{P}(\overline{S} \leq Q_{2\epsilon}(\overline{S}))} \\
&\overset{(c)}\leq \frac{\sigma_S}{\sqrt{2\epsilon}}2\epsilon + \sigma_S\sqrt{\mathbb{P}(\overline{S} \leq Q_{2\epsilon}(\overline{S}))} \\
&= \sigma_S\sqrt{8\epsilon},
\end{align*}
where (a) applies the Cauchy–Schwarz inequality, and (b) follows from the definition and upper bound of quantile in \eqref{eqn: def Q} and \eqref{eqn: upper bound on Q}, and (c) is given by $\epsilon < 0.25$ by the assumption $t_0 > \frac{12\log(4/\delta)}{0.25 - 8\eta}$. 
The proof is similar for the second term on the right-hand side of \eqref{eqn: bias 1}, and we obtain 
\begin{equation} \label{eqn: trim exp and true mean 1}
\left|\mathbb{E}[\phi_{t_0}(S)] - S^* \right| \leq \sigma_S\sqrt{8\epsilon}.
\end{equation}

Using \eqref{eqn: beta} and \eqref{eqn: alpha}, and the upper bound on $\mathbb{E}[\phi_{t_0}(S)]$ derived from \eqref{eqn: trim exp and true mean 1}, we have that any data point $s_k$,
\begin{align}\label{eqn: trim est and trim mean 1}
    &\left|\phi_{t_0}(s_k)-\mathbb{E}[\phi_{t_0}(S)]\right| \nonumber\\
    &\quad \leq \max\{|Q_{\epsilon/2}(\overline{S})|, |Q_{1-\epsilon/2}(\overline{S})|\} + \sigma_S\sqrt{8\epsilon} \nonumber\\
    &\quad \leq \frac{\sigma_S}{\sqrt{\epsilon/2}} + \sigma_S\sqrt{8\epsilon},
\end{align} where the last inequality follows from \eqref{eqn: upper bound on Q} and $\epsilon < 0.25$.

Applying Bernstein's inequality in Lemma \ref{lemma: bern}, where $a$ is the right-hand side of \eqref{eqn: trim est and trim mean 1}, with probability at least $1-\delta/2$, we bound the second term of \eqref{eqn: estimate and true 1} as
\begin{multline}\label{eqn: trim est and trim mean bern 1} 
\left|\frac{1}{t}\sum_{k = 1}^t \phi_{t_0}({s}_k)-\mathbb{E}[\phi_{t_0}(S)]\right| 
    \leq \sigma_S \sqrt{\frac{2\log(4/\delta)}{t}} \\
    + \frac{2\sigma_S}{\sqrt{\epsilon/2}} \frac{\log(4/\delta)}{3t}
    + \frac{2\sigma_S\sqrt{8\epsilon}\log(4/\delta)}{3t}.
\end{multline}  

Expanding the second term of on the right-hand-side of \eqref{eqn: trim est and trim mean bern 1}, we obtain 
\begin{multline}\label{eqn: Q log upper 1}
   \frac{2\sigma_S}{\sqrt{\epsilon/2}} \frac{\log(4/\delta)}{3t} 
    \overset{(a)}\leq \frac{2\sigma_S\log(4/\delta)}{3t\sqrt{4\eta + 6\frac{\log(4/\delta)}{t_0}}} \\
    \overset{(b)}\leq \sigma_S\frac{2\sqrt{t_0}}{3t}\sqrt{\frac{\log(4/\delta)}{6}} 
    \overset{(c)}\leq \sigma_S \frac{2}{3}\sqrt{\frac{\log(4/\delta)}{6t}},
\end{multline} where (a) and (b) are derived from the definition of $\epsilon$ in \eqref{eqn: epsilon def}, and (c) follows from $t_0/t \leq 1, \forall t \geq t_0$.
Noting that $\log(4/\delta)/t \leq 1$ by the assumption that $\delta \geq 4e^{-t_0}$, we have
\begin{multline}\label{eqn: trim est and trim mean final 1} 
\left|\frac{1}{t}\sum_{k = 1}^t \phi_{t_0}({s}_k)-\mathbb{E}[\phi_{t_0}(S)]\right| \\
    \leq \left(\sqrt{2} + \frac{\sqrt{6}}{9}\right)\sigma_S \sqrt{\frac{\log(4/\delta)}{t}} + {\frac{4\sqrt{2}}{3}}\sigma_S\sqrt{\epsilon}.
\end{multline}

For corrupted data satisfying $\phi_{t_0}(\tilde{s}_t) \neq \phi_{t_0}({s}_t)$ at time $t$, the gap is bounded 
\begin{equation*}
    \left|\phi_{t_0}(\tilde{s}_t) - \phi_{t_0}({s}_t) \right| \leq |Q_{\epsilon/2}(\overline{S})| + |Q_{1-\epsilon/2}(\overline{S})|,
\end{equation*}
and it follows that,
\begin{align} \label{eqn: thm1 corruption}
&\frac{1}{t}\Big|\sum_{k = 1}^t \phi_{t_0}(s_k) - \sum_{k = 1}^t \phi_{t_0}(\tilde{s}_k)\Big| \nonumber \\
&\quad \overset{(a)}\leq \eta \left(|Q_{\epsilon/2}(\overline{S})| + |Q_{1-\epsilon/2}(\overline{S})|\right) \nonumber \\
&\quad \overset{(b)}\leq \eta \left(\frac{\sigma_S}{\sqrt{1-\epsilon/2}}+\frac{\sigma_S}{\sqrt{\epsilon/2}}\right) \nonumber\\
&\quad \overset{(c)}\leq \frac{\epsilon}{8} \left(\frac{2\sigma_S}{\sqrt{\epsilon/2}}\right)
\leq \frac{\sigma_S\sqrt{2\epsilon}}{4},
\end{align}
where (a) reflects the potential presence of up to $\eta t$ corrupted samples at time $t$, (b) follows from \eqref{eqn: upper bound on Q}, and (c) follows from $\eta \leq \epsilon/8$ by definition in \eqref{eqn: epsilon def} and $\epsilon< 0.25$.

Using \eqref{eqn: trim exp and true mean 1}, \eqref{eqn: trim est and trim mean final 1}, and \eqref{eqn: thm1 corruption} in \eqref{eqn: estimate and true 1}, with probability at least $1-\delta$,
\begin{equation}\label{eqn: algo 1 error bound complete}
|\hat{S}_{t} - S^*| \leq \left(\sqrt{2} + \frac{\sqrt{6}}{9}\right) \sigma_S \sqrt{\frac{\log(4/\delta)}{t}}
+ \frac{43\sqrt{2}}{12}\sigma_S\sqrt{\epsilon}.
\end{equation}

\subsection{Proof of Corollary \ref{cor: sample cov asymp bound}}
Let $\rho_{ij,t} = \left(\sqrt{2} + \frac{\sqrt{6}}{9}\right) \sigma_{S_{ij}} \sqrt{\frac{\log(4/\delta)}{t}} + \frac{43\sqrt{2}}{12} \sigma_{S_{ij}}\sqrt{\epsilon}$.
From Theorem \ref{thm: online 1}, we have
\begin{equation}
    \mathbb{P}[|\hat{S}_{ij,t}-S_{ij}^*| \geq {\rho}_{ij,t}] \leq \delta_t.
\end{equation}
Applying the union bound, we have  
\begin{equation}\label{eqn: badeventprob}
    \mathbb{P}[\cup_{ij} \{|\hat{S}_{ij,t}-S_{ij}^*| \geq {\rho}_{ij,t}\}] \leq p^2\delta_t.
\end{equation}
This result holds for any fixed $t$. Note that from Theorem \ref{thm: online 1}, the range for the selection of $\delta$ is time-dependent. We use $\delta_t$ to represent this dependency and each $\delta_t \in (4e^{-t_0},1/p^2)$. We will now extend this to bound the deviation of $\hat{S}_t$ from $S^*$ \textit{for all} sufficiently large $t$.

Define a bad event at time $t$ as the event that there exists a pair of $(i,j)$ such that $|\hat{S}_{t,ij}-S^*_{ij}| > \rho_{ij,t}$. Define the random variable $B_t$, with $B_t = 1$ if the bad event occurs at the given $t$, and $0$ otherwise. Define $B = \sum_{k = t_0}^t B_k$ as the number of bad events up to time $t$. We can always select a constant $c$, such that $\delta_t = ce^{-t} \in (4e^{-t_0},1/p^2)$. Summing up the probability of bad events in \eqref{eqn: badeventprob}, we have 
\begin{align*}
    \sum_{k = t_0}^t \mathbb{P}[B_k]
    &\leq cp^2 \sum_{k=t_0}^t e^{-t} < \infty.
\end{align*}

From the Borel-Cantelli lemma, the probability of infinitely many bad events occurring is 0. Thus, for a set of sample paths of measure 1, there exists a sample-path dependent finite time $\bar{t}$ such that for $t \geq \bar{t}$, no bad event occurs.
In other words, for all $t \geq \bar{t}$, $|\hat{S}_{ij,t} - S^*_{ij}| \leq \rho_{ij,t}$ $\forall i,j$ almost surely, and $\max_{ij}(|\hat{S}_{t,ij}-S^*_{ij}|)\leq \max_{ij}\rho_{ij,t}$. 
Using the matrix norm inequality, $\|\hat{S}_t-S^*\|_F \leq p\rho_t$, where $\rho_t = \max_{ij} \rho_{ij,t}$. 
Thus, for each sample path in a set of measure 1, there exists a finite time $\bar{t}$ such that $\forall t\geq \bar{t}$,
\begin{equation}
    \|\hat{S}_t - S^*\|_F \leq p\rho_t = p\max_{ij}\rho_{ij,t}.
\end{equation}

\subsection{Proof of Theorem \ref{theorem: eigen}}
Consider the update for $\Gamma_t$ given by \eqref{eqn: dual update}.
According to Weyl's inequality and the upper bound of the eigenvalues of the clipping operator (see \cite{OGAMA}), the eigenvalues of $\Gamma_{t}$ for fixed $t$ satisfy
\begin{align*}
    \lambda_{\max}(\Gamma_{t})     
    &\leq p\lambda+\lambda_{\max}(\hat{S}_t).
\end{align*}

A symmetric matrix with positive diagonal elements has at least one positive eigenvalue, thus, $\lambda_{\max}(\hat{S}_t) > 0, \forall t$. Define $b = p\lambda + \max_{t \ge t_0}\lambda_{\max}(\hat{S}_t)$. We have $0 < \lambda_{\max}(\Gamma_t) \leq b$ for all finite $t$.  

Now we show the lower bound for eigenvalues.
The step size $\zeta_t$ is chosen to guarantee convergence as shown in Theorem \ref{theorem1}. More specifically, it can be chosen for each iteration by backtracking line search such that for the next iteration, $\Gamma_{t}$ satisfies the sufficient descent condition 
\begin{equation}
    f(\Gamma_{t}) \leq D_{\zeta_{t-1}}(\Gamma_{t},\Gamma_{t-1}),
\end{equation} 
where $f(\Gamma_t) = -\log\det\Gamma_t$ is the dual objective from \eqref{eqn: dual} and $D_{\zeta_{t-1}}$ is the quadratic approximation of the dual objective \eqref{eqn: dual} around $\Gamma_{t-1}$ given by
\begin{multline}
    D_{\zeta_{t-1}}(\Gamma_{t},\Gamma_{t-1}) = f(\Gamma_{t-1}) + \trace[(\Gamma_{t}-\Gamma_{t-1})\nabla f(\Gamma_{t-1})] 
    \\ + \frac{1}{2\zeta_{t-1}}\|\Gamma_{t}-\Gamma_{t-1}\|_F^2.
\end{multline}
Using the approximation and computing the derivative $\nabla f$, we have the sufficient descent condition 
\begin{equation}\label{eqn: logdet bound}
    -\log \det \Gamma_{t} \leq -\log \det \Gamma_{t-1} + \Delta_{t-1}, 
\end{equation} where $\Delta_{t} = \trace[\left(\Gamma_{t+1}-\Gamma_{t}\right)(-\Omega_{t})]+\frac{1}{2\zeta_{t}}\|\Gamma_{t+1}-\Gamma_{t}\|_F^2$.

The step size can be selected small enough to satisfy the above conditions, and it was shown in \cite{dalal2017sparse} that $\zeta_t \leq \min(\lambda_{\min}(\Gamma_{t-1}, \Gamma_{t}))^2$ is sufficient to satisfy the condition.

Iterating through the above condition, we obtain 
\begin{equation}
    \log \det \Gamma_{t} \geq \log \det \Gamma_{t_0}  - \sum_{k =t_0}^{t-1} \Delta_k.
\end{equation}

Let $a_{t}$ be the smallest eigenvalue of $\Gamma_{t}$, and let $g(t) = \log \det \Gamma_{t_0}  - \sum_{k =t_0}^{t-1} \Delta_k$, we have 
\begin{align}
    \log a_t + (p-1)\log b  &\geq \log \det \Gamma_{t} \geq g(t)\\
    a_{t+1} &\geq e^{g(t)} b^{1-p}.
\end{align}
For a finite time step $t \geq t_0$, we can always select a large enough $\lambda$, such that $g(t) > - \infty$ for all $t_0 \leq \ldots \leq t$ and $a_{t} \geq a$, where $a=\min_{t_0 \leq t}(a_t)$.

Thus, we have $aI_p \preceq \Gamma_t \preceq bI_p$ for all finite $t \ge t_0$.

\subsection{Proof of Lemma \ref{lemma:nedic}}
Let $\xi>0$ be arbitrary. Since $\lim_{t\to\infty}\rho_t \leq \bar{\rho}$, we let $\bar{t}$ to be the index such that $\rho_t \leq \bar{\rho} + \xi$ for all $t \geq \bar{t}$.

For all $t \geq \bar{t} + 1$, we can partition
\begin{equation}
    \sum_{l=0}^{t} r^{t-l}\rho_l = \sum_{l=0}^{\bar{t}} r^{t-l}\rho_l + \sum_{l=\bar{t}+1}^{t} r^{t-l}\rho_l.
\end{equation}
Using properties of geometric series, we have
\begin{align}
    \sum_{l=0}^{\bar{t}} r^{t-l}\rho_l 
    \leq \max_{0\leq k \leq \bar{t}} \rho_k \sum_{l=0}^{\bar{t}} r^{t-l} 
    &= \max_{0\leq k \leq \bar{t}} \rho_k r^{t-\bar{t}}\sum_{t=0}^{\bar{t}} r^{t} \nonumber\\
    & \leq \max_{0\leq k \leq \bar{t}} \rho_k \frac{r^{t-\bar{t}}}{1-r},
\end{align}
and 
\begin{equation}
    \sum_{l=\bar{t}+1}^{t} r^{t-l}\rho_l \leq (\bar{\rho}+\xi)\sum_{l=\bar{t}+1}^{t} r^{t-l}  \leq \frac{\bar{\rho}+\xi}{1-r}.
\end{equation}
Therefore, by combining the results, we have 
\begin{equation}
    \sum_{l=0}^{t} r^{t-l}\rho_l \leq \max_{0\leq k \leq \bar{t}} \rho_k \frac{r^{t-\bar{t}}}{1-r} + \frac{\bar{\rho}+\xi}{1-r}.
\end{equation}
Since $\xi$ is arbitrary and $r\in(0,1)$, we have the result 
\begin{equation}
    \limsup_{t\to\infty} \sum_{l=0}^{t} r^{t-l}\rho_l \leq \frac{\bar{\rho}}{1-r}.
\end{equation}

\subsection{Proof of Corollary \ref{cor: convergence}}
From Theorem \ref{theorem1}, if $\forall t\ge t_0$, $\zeta_t^k = \zeta < a^2$, then $0 < r < 1$. The first term $r^{t+1-t_0}\|\Gamma_{t_0}-\Gamma^*\|_F$ in \eqref{eqn:bound_gamma_t} converges exponentially to zero. The second term in \eqref{eqn:bound_gamma_t} is an instance of Lemma \ref{lemma:nedic}, with $r < 1$ and $\rho_t \to \bar{\rho}$ almost surely as $t\to \infty$ from Corollary \ref{coro: asymp cov}.

\subsection{Boundedness of Eigenvalues for Sparse Cases}
\begin{figure}
    \centering
    \includegraphics[width = 0.8\linewidth]{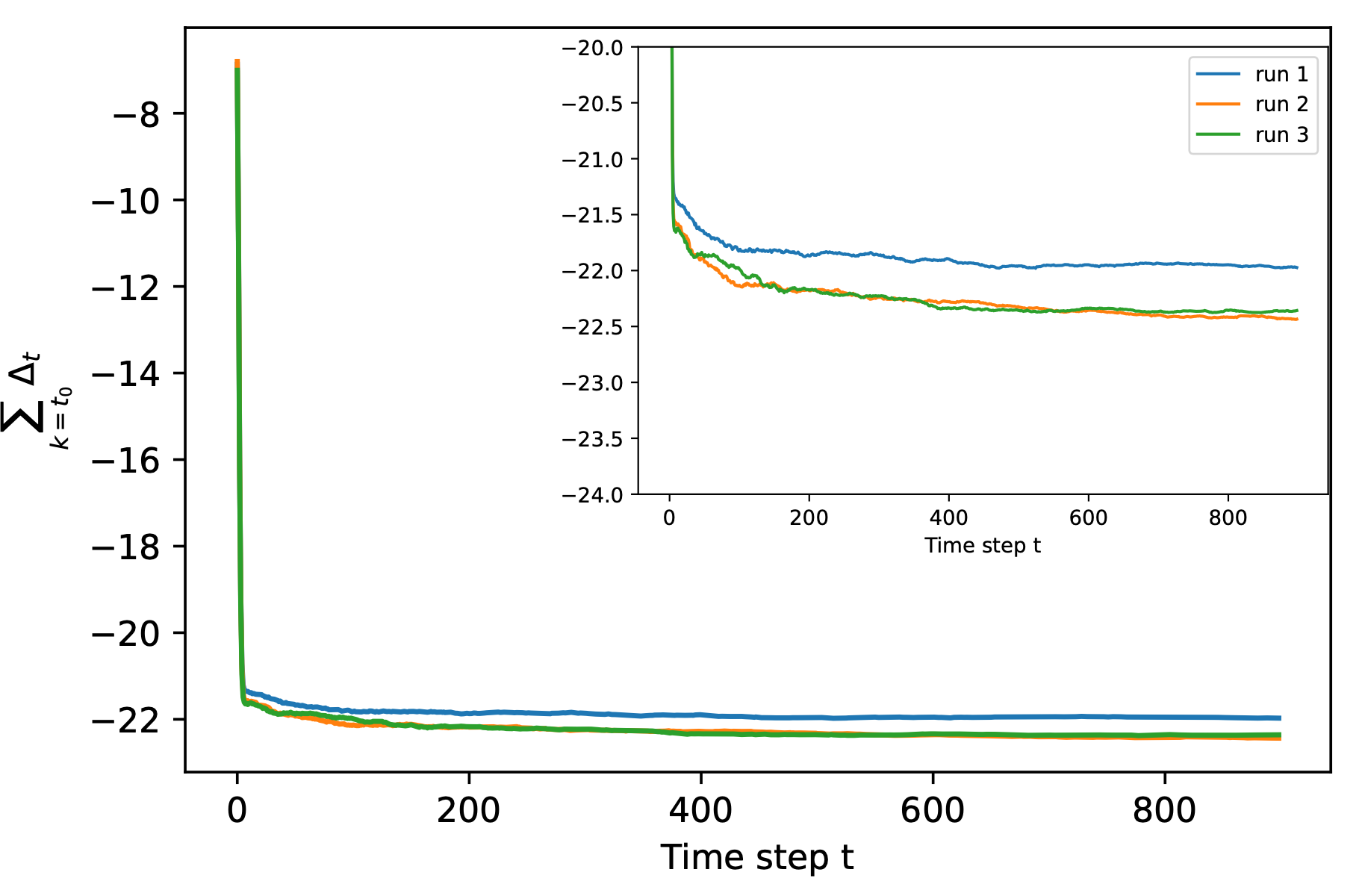}
    \caption{Boundedness of the sum of error terms $\Delta$.}
    \label{fig: error sum}
\end{figure}

In Fig. \ref{fig: error sum}, we demonstrate the boundedness of error terms defined in \eqref{eqn: logdet bound}, by running 3 independent experiments where $p >> t_0$. The subplot with an adjusted $y$-axis shows the values with a higher resolution. The total number of data points is 1000, $t_0 = 100$, $p = 500$, $\lambda = 0.5$, and other parameters are identical to the previous experiment setup. 

\end{document}